\documentclass[10pt,twocolumn,letterpaper]{article}

\usepackage[pagenumbers]{wacv} 
\usepackage{multirow}
\usepackage{float}

\definecolor{wacvblue}{rgb}{0.21,0.49,0.74}
\usepackage[pagebackref,breaklinks,colorlinks,allcolors=wacvblue]{hyperref}

\def\wacvPaperID{*****} 
\def\confName{WACV}
\def\confYear{2026}

\title{FSP-DETR: Few-Shot Prototypical Parasitic Ova Detection}

\author{Shubham Trehan$^1$, Udhav Ramachandran$^2$, Akash Rao$^1$, Ruth Scimeca$^2$, \\Sathyanarayanan N. Aakur$^1$\\
$^1$CSSE Department, Auburn University, Auburn, AL, 36849\\
$^2$Department of Veterinary Pathobiology, Oklahoma State University, Stillwater, OK, 74078\\
{\tt\small \{szt0113,azr0187,san0028\}@auburn.edu}, \tt\small \{udhav.ramachandran,ruth.scimeca\}@okstate.edu
}

\begin{document}
\maketitle

\begin{abstract}
Object detection in biomedical settings is fundamentally constrained by the scarcity of labeled data and the frequent emergence of novel or rare categories. We present FSP-DETR, a unified detection framework that enables robust few-shot detection, open-set recognition, and generalization to unseen biomedical tasks within a single model. Built upon a class-agnostic DETR backbone, our approach constructs class prototypes from original support images and learns an embedding space using augmented views and a lightweight transformer decoder. Training jointly optimizes a prototype matching loss, an alignment-based separation loss, and a KL divergence regularization to improve discriminative feature learning and calibration under scarce supervision. Unlike prior work that tackles these tasks in isolation, FSP-DETR enables inference-time flexibility to support unseen class recognition, background rejection, and cross-task adaptation without retraining. We also introduce a new ova species detection benchmark with 20 parasite classes and establish standardized evaluation protocols. Extensive experiments across ova, blood cell, and malaria detection tasks demonstrate that FSP-DETR significantly outperforms prior few-shot and prototype-based detectors, especially in low-shot and open-set scenarios.
\end{abstract}

\section{Introduction}\label{sec:intro}

Automated visual understanding in biomedical imaging is essential for advancing diagnostics, epidemiology, and large-scale disease surveillance. Yet, in many real-world scenarios, such as identifying rare parasitic infections, novel cell morphologies, or emerging pathogens, data (both annotated and raw) is scarce or entirely unavailable. This presents a unique challenge: models must often learn from just a few labeled examples, contend with the possibility of encountering previously unseen categories, and generalize to new domains without exhaustive retraining. Addressing this reality requires systems that support few-shot detection to operate in low-label regimes, enable open-set detection to flag unfamiliar or novel instances, and achieve zero-shot generalization to extend recognition capabilities beyond the training distribution, all while maintaining high precision and minimizing false alarms in sensitive clinical workflows.

\begin{figure}
    \centering
    \includegraphics[width=0.99\linewidth]{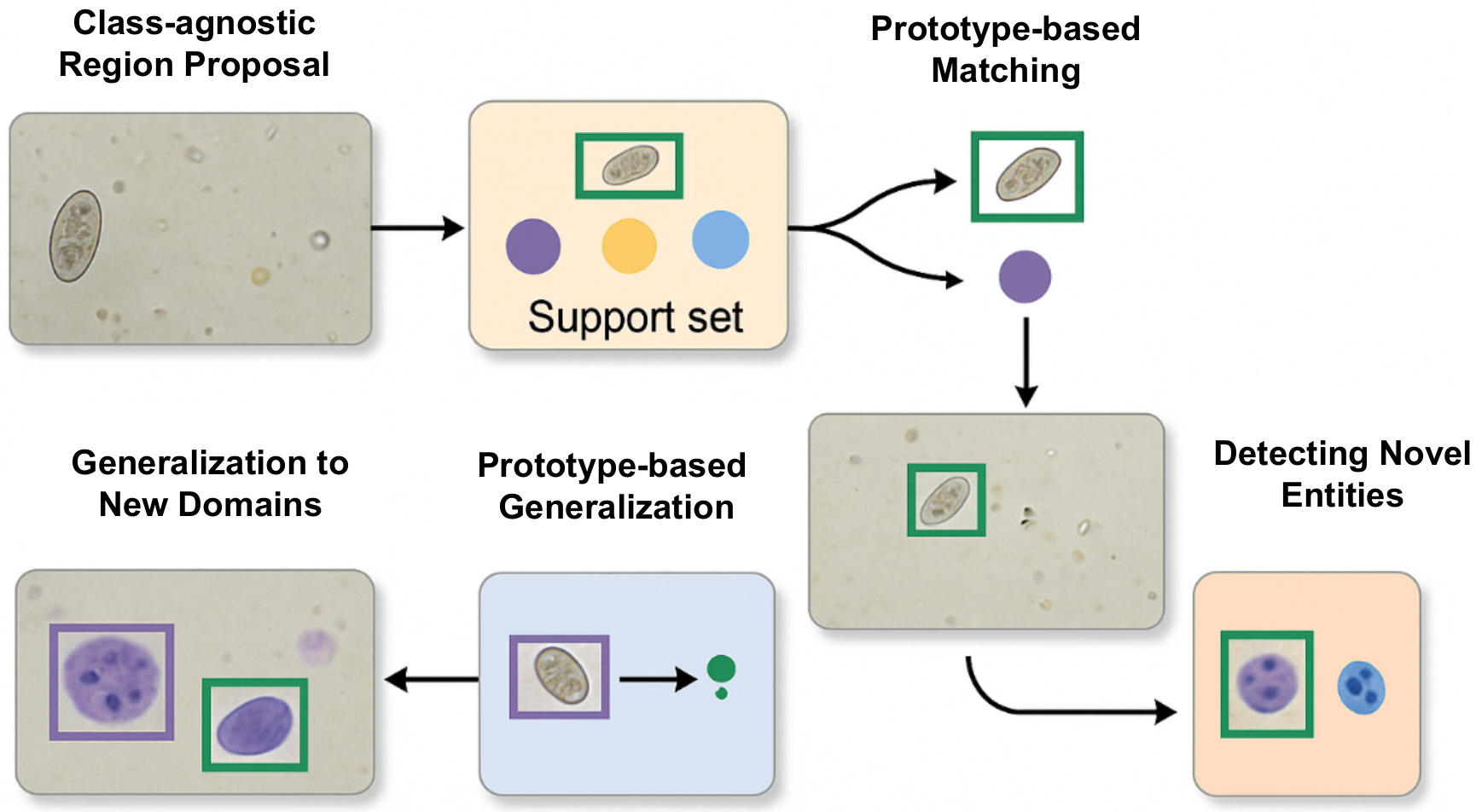}
    \caption{\textbf{Overview.} Our proposed FSP-DETR framework provides a simple, modular unified framework for robust few-shot detection from limited support examples, detection of novel/unseen entities through open-set reasoning, and cross-task transfer to new biomedical domains without extensive tuning.} 
    \label{fig:enter-label}
\end{figure}

While few-shot \cite{Kohler2023FewShotSurvey,Xin2024FewShotAdvances,Fan2020AttentionRPN}, open-set \cite{Zheng2022OpenSetDiscovery,Ammar2024OpenSetBenchmarking,Liu2024GroundingDINO}, and zero-shot \cite{Bansal2018ZeroShot,Ma2025FineGrainedZSD,Jayasekara2022UnifiedFramework} recognition have each received growing attention, they are typically addressed in isolation and rarely within a unified object detection framework. 
In biomedical imaging, however, these challenges frequently co-occur: models must not only detect and classify known structures with limited supervision but also identify novel entities or generalize across diagnostic contexts. 
Addressing them separately often leads to brittle pipelines and fragmented solutions. To this end, we propose a simple yet powerful energy-based detection framework that jointly models these capabilities. Our approach is motivated by the insight that class prototypes, constructed from very few examples, can serve as a flexible representation to support detection, classification, and novelty rejection. 

\begin{figure*}
    \centering
    \includegraphics[width=0.95\textwidth]{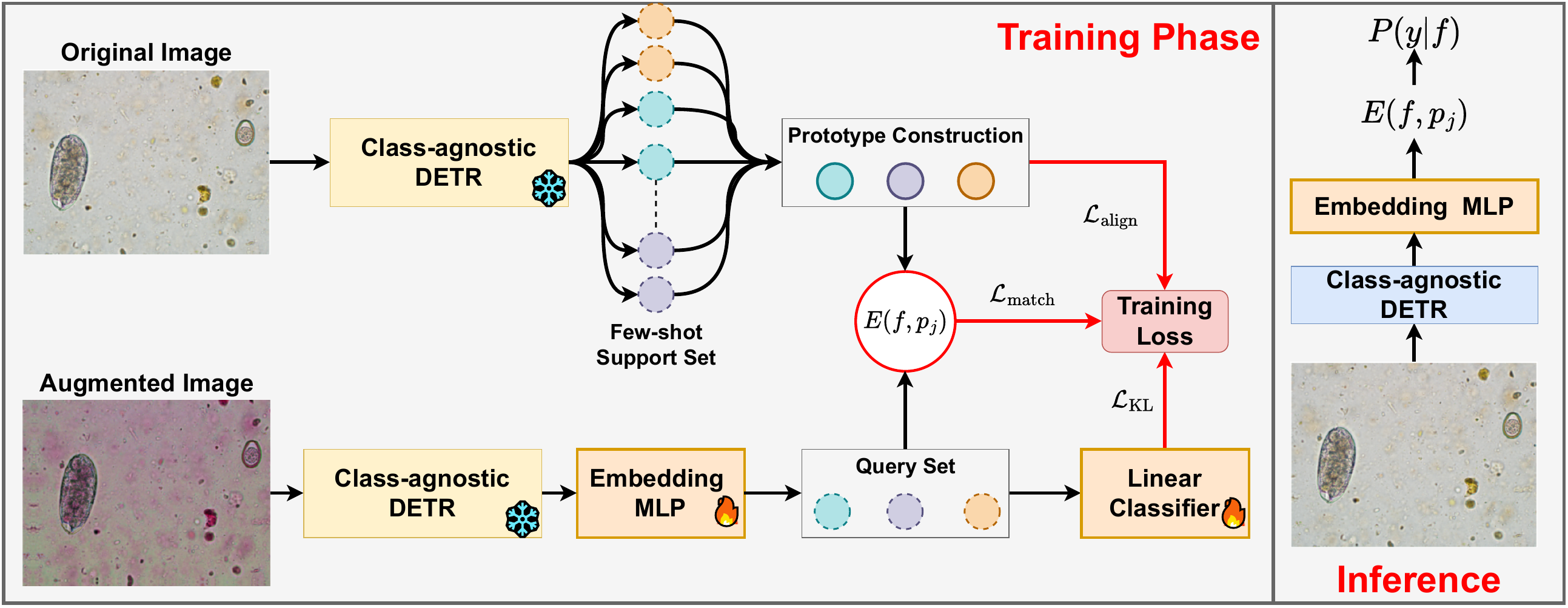}
    \caption{\textbf{Framework Overview.}  
    The proposed method uses a class-agnostic DETR backbone to extract object proposals from both original and augmented images. 
    Features from the original images are used to construct class prototypes, while augmented image features are passed through a trainable embedding module and classifier. 
    The training objective jointly optimizes prototype alignment and refinement losses. 
    At inference, the model can recognize seen and novel categories by distance-based classification with learned prototypes.}
    \label{fig:pipeline}
\end{figure*}

Our proposed framework, FSP-DETR (Figure~\ref{fig:pipeline}), extends the foundational idea of prototypical networks (ProtoNets)~\cite{snell2017prototypical}, which represent each class by a prototype in embedding space and classify instances via distance-based matching, to the more complex setting of object detection. 
Unlike standard ProtoNets, which operate on whole-image classification, our method introduces a modular two-stage architecture. 
First, a class-agnostic DETR~\cite{carion2020end} backbone is trained solely to detect ova-like regions in a binary ova-vs-background setting, without relying on subtype labels, allowing the model to learn transferable morphological priors. 
Second, we perform prototype-based inference using support examples to recognize fine-grained subtypes from a small labeled pool. 
We further draw from ProtoKD~\cite{trehan2023protokd} and extend it in two key directions: (i) adapting it to detection via transformer-based region proposal features, and (ii) jointly optimizing prototype alignment, distance-based separation, and KL-regularized classification to enhance cross-class generalization. 
This decoupled formulation not only improves interpretability and scalability but also facilitates few-shot learning, open-set recognition, and cross-domain transfer without architectural changes. Extensive experiments across ova, blood cell, and malaria detection tasks validate our approach, showing consistent gains in low-shot and zero-shot settings with a compact and modular design.

Our \textbf{contributions} are fourfold. We 
(i) present a unified detection framework for few-shot learning, open-set recognition, and cross-task generalization via class-agnostic proposals and prototype-based inference, 
(ii) introduce a training strategy that jointly optimizes feature alignment, distance-based separation, and classifier regularization to enhance discrimination under limited supervision, 
(iii) decouple prototype construction and embedding learning using original and augmented views to improve representation robustness, and 
(iv) curate and publicly release an ova detection dataset with bounding box annotations across 20 parasite classes (including seen/unseen splits), enabling systematic low-shot evaluation. 

\section{Related Work}\label{sec:rw}

\textbf{Few-Shot Detection and Meta-Learning.}  
Meta-learning has become a cornerstone of few-shot object detection, focusing on rapid adaptation with minimal examples through episodic training. Meta-DETR~\cite{zhang2022meta} avoids explicit region proposals by leveraging class correlations, while Meta Faster R-CNN~\cite{han2022meta} addresses noisy proposals and spatial misalignment via attentive feature alignment and prototype-based heads. Wang et al.~\cite{wang2020few} showed that fine-tuning simple architectures can outperform complex meta-learners, underscoring the power of straightforward approaches. Complementary methods introduce feature reweighting and attention to enhance generalization: Kang et al.~\cite{kang2019few} propose a meta feature learner for dynamic scaling, FSCE~\cite{sun2021fsce} uses contrastive proposal encoding for class separation, and Fan et al.~\cite{fan2020few} embed attention into both proposal and classification stages. Further innovations include semantic graphs~\cite{zhu2021semantic}, hallucinated RoI features~\cite{Zhang_2021_CVPR}, and cross-transformers~\cite{han2022few}, all of which refine few-shot detection under data scarcity. 
Recent detection models like FCOS~\cite{tian2019fcos} and GFocal~\cite{fei2025gfocal} offer dense, anchor-free architectures that can aid few-shot detection through better localization priors, though they lack mechanisms for prototype-based adaptation. Stable-DINO’s~\cite{Liu2024GroundingDINO} transformer-based~\cite{vaswani2017attention} design shows strong performance, but struggles to generalize under few-shot constraints without extensive tuning. GeCo~\cite{pelhan2024novel} highlights the potential of prototype inference in low-shot settings, but is limited to single-class counting and lacks open-set or multi-class support. 

\textbf{Metric Learning and Transfer.}  
Prototype-based methods reformulate few-shot detection as a metric learning problem by enabling distance-based classification in an embedding space. RepMet~\cite{karlinsky2019repmet} jointly learns embeddings and multimodal prototypes, enabling robust comparison without traditional parametric classifiers. FSCE~\cite{sun2021fsce} enhances this with supervised contrastive losses on object proposals, and Meta Faster R-CNN~\cite{han2022meta} improves alignment through hierarchical prototype matching. Beyond metric learning, transfer-based methods address data limitations via synthetic augmentation and large-scale pretraining. For instance, synthetic feature generation using optimal transport aligns novel-class distributions with base data, and LSTD~\cite{Han_2024_CVPR} fuses SSD and Faster R-CNN for structured knowledge transfer. Emerging foundation model approaches leverage contrastive pretraining and vision-language alignment, using LLM-derived relational context to classify novel objects with minimal supervision.

\textbf{Biomedical Detection and Ova Classification.}  
In biomedical domains, deep learning increasingly enables fine-grained parasite and blood cell recognition. He et al.~\cite{He2023ResTFG} combine CNNs and transformers for accurate \textit{Eimeria} classification in poultry, while Ray et al.~\cite{Ray2021AscarisNecator} apply robust ensembles for identifying ova such as \textit{Ascaris lumbricoides} and \textit{Necator americanus} in swine diagnostics. Weakly supervised learning frameworks have also been employed in malaria and sickle cell detection~\cite{Manescu2020WeaklyMalariaSickle}, reducing annotation effort while maintaining diagnostic accuracy. These approaches underscore the importance of scalable, low-supervision methods in biomedical image analysis, particularly relevant for our proposed framework, which targets few-shot, open-set, and zero-shot ova detection in low-resource clinical settings. 

\section{Our Approach}\label{sec:methods}

\textbf{Overview.} 
We address the task of few-shot open-vocabulary ova detection and classification in microscopic images. Given a training set with only a few annotated examples per class, our goal is to develop a system that can (i) localize potential ova instances in unseen images, (ii) classify them into known categories using learned prototypes, and (iii) reject detections that do not correspond to any known class. 
We postulate that this task can be effectively addressed by learning a discriminative embedding function in which classification is performed via distance-based comparison to a small set of representative examples. If bounding boxes can be mapped into a feature space where instances of the same class are close together and instances of different classes are far apart, then classification can be achieved by assigning each query to its nearest prototype. More importantly, this embedding must generalize from sparse supervision and support open-set rejection. 
We formalize this using an energy-based perspective, where the \emph{energy} between a query embedding and a class prototype reflects their semantic compatibility. The correct class should have low energy (i.e., small distance in feature space), while incorrect or unknown classes should have high energy. Let $f \in \mathbb{R}^d$ denote the feature embedding of a candidate bounding box, and let $\{p_j\}_{j=0}^C$ be the set of prototypes, where $j = 0$ denotes the background class. The probability of class membership is defined as:

\begin{equation}
\label{eqn:probs}
P(y = j \mid f) = \frac{\exp(-E(f, p_j))}{\sum_{k=0}^C \exp(-E(f, p_k))}
\end{equation}

\noindent where $E(f, p_j) = d(f, p_j)$ measures the distance between the query and the prototype. 
To train the model, we minimize a structured energy objective that combines four terms: a class-agnostic detection loss to localize candidate regions, a prototype matching loss to minimize energy to the correct class, a distillation loss to align classifier predictions with prototype-based scores, and a soft alignment-based loss to enforce inter-class separation. Formally, we define it as

\begin{equation}
\label{eq:energy}
\min_{\theta, \phi} \sum_{b \in \mathcal{D}_\theta(I)} \left[
E_{\text{box}} + E_{\text{match}} + 
 D_{\text{KL}} + 
E_{\text{align}} \right]
\end{equation}

This energy can be minimized by learning an embedding space in which the distance between each query and its corresponding class prototype is minimized, while distances to other class prototypes are maximized. 

\subsection{Class-Agnostic Detection Backbone}
\label{sec:detection}

We begin by identifying candidate ova regions using a class-agnostic object detector. Specifically, we adopt a DETR-based architecture~\cite{carion2020end}, which we fine-tune in a binary detection setting, treating all annotated ova instances as a single foreground class. This formulation enables the detector to focus solely on learning spatial and morphological cues shared across ova types, without relying on subtype-specific supervision. Background regions are implicitly handled via DETR's bipartite matching mechanism, which allows for precise localization even under limited annotation. 
Formally, for an input image $I$, the detector $\mathcal{D}\theta$ outputs a set of bounding boxes $\mathcal{B} = {b_1, \dots, b_K}$ and corresponding visual features $v_k \in \mathbb{R}^d$ for each $b_k$. These features are derived from the encoder-decoder representations and serve as input to the downstream prototype-based classifier. The detector is optimized using DETR’s standard loss ($E{\text{box}}$), which combines localization and objectness terms to guide learning of region-level proposals. 
By training only to detect ``ova-like'' regions, the backbone learns transferable visual priors that support generalization without encoding class semantics, a design that remains faithful to the few-shot setting. Since subtype recognition is deferred to a separate prototype-based module trained solely on limited support data, our framework maintains a clean separation between class-agnostic localization and low-shot classification. While this setup may yield noisy proposals under low-resource conditions, it allows for robust downstream reasoning over localized candidates and supports scalability to novel or unseen categories.

\subsection{Prototype Construction and Query Generation}
\label{sec:prototypes}

With a set of candidate bounding boxes ($\mathcal{B}$) from the class-agnostic detector, we next construct a set of prototypes ($\{p_j\}_{j=1}^C$) to represent each class, along with queries that simulate test-time variation. These form the basis for learning a discriminative embedding space for few-shot classification. 
We begin by sampling a small number of labeled exemplars per class from the training set. For each exemplar, we extract a feature vector ($v_i$) from the DETR decoder and pass it through a lightweight two-layer MLP $\phi(\cdot)$ to obtain an embedding. Class prototypes are then formed by averaging the resulting embeddings across support examples:

\begin{equation}
p_j = \frac{1}{n} \sum_{i=1}^{n} \phi(v_i^{(j)}), \quad j \in \{1, \dots, C\}
\end{equation}

To support open-set recognition, we also include a background prototype $p_0$, computed by averaging features from predicted boxes with low overlap ($IoU < 0.3$) with any ground truth instance. This allows the model to learn a notion of “none-of-the-above” and reject spurious detections.

To generalize beyond the few available exemplars, we construct a set of queries by applying randomized photometric and geometric transformations to the support images. These include flips, rotations, color jitter, and additive noise, chosen to reflect natural variation in microscope imaging. The resulting augmented images are processed through the same DETR model, and features are extracted from the corresponding boxes to form queries. This helps simulate potential variability likely to be encountered at test time and provides regularization to encourage the embedding space to remain stable under appearance shifts. 

\subsection{Embedding Alignment via Prototype Matching}
\label{sec:emb_alignment}

The first stage of training builds on Prototypical Networks~\cite{snell2017prototypical}, which aim to learn an embedding space where features from the same class are pulled close to a shared prototype, while being pushed away from others. Given the prototypes $\{p_j\}_{j=1}^C$ constructed from support examples and the query embeddings $\{q_i\}$ derived from augmented inputs, we define a matching-based objective that encourages alignment between each query and its corresponding class prototype. 
We define the energy between a query embedding $q_i$ and a prototype $p_j$ as the squared Euclidean distance:
\[
E(q_i, p_j) = \| q_i - p_j \|^2
\]
The class probabilities are then computed via a softmax over negative energies as defined in Equation~\ref{eqn:probs}. 
The prototype matching loss minimizes the negative log-likelihood of the ground-truth class:
\begin{equation}
\mathcal{L}_{\text{match}} = - \sum_i \log P(y_i \mid q_i)
\end{equation}

\noindent and encourages the embedding network $\phi(\cdot)$ to project augmented queries close to their respective prototypes in feature space. Because queries are derived from perturbed versions of the support set, this process also implicitly regularizes the embedding by promoting invariance to intra-class variation.  However, while encouraging queries to move closer to their corresponding prototypes, it does not explicitly discourage proximity to incorrect classes. In settings with limited supervision, this can result in overlapping or poorly separated class clusters, reducing classification robustness. To more effectively shape the embedding space, we introduce additional objectives that promote inter-class separation and alignment with a learned classifier, thereby improving discriminability under data-scarce conditions.

\subsection{Embedding Refinement}
\label{sec:emb_refinement}

To improve class separability and reinforce discriminative structure in the embedding space, we introduce two complementary objectives: a distillation loss that aligns the outputs of a learned classifier with prototype-based predictions, and a metric learning loss that enforces separation between different class regions. 
In addition to prototype-based classification, we train a shallow linear classifier on top of the learned embeddings. To ensure consistency between the prototype-based predictions and the classifier outputs, we apply a Kullback–Leibler (KL) divergence loss that encourages the two distributions to match, given by 

\begin{equation}
\mathcal{L}_{\text{KL}} = \sum_i D_{\text{KL}} \left( P_{\text{proto}}(y \mid q_i) \,\|\, P_{\text{clf}}(y \mid q_i) \right)
\end{equation}
where \( P_{\text{proto}}(y \mid q_i) \) denotes the softmax over distances to prototypes (as defined earlier), and \( P_{\text{clf}}(y \mid q_i) \) is the classifier’s prediction. 
This encourages the classifier to reflect the structure learned by the prototype matching objective while offering additional flexibility during inference. 
To encourage inter-class separation while maintaining robust alignment with class prototypes, we employ a \textit{temperature-scaled similarity-based loss} inspired by contrastive learning. Building on the InfoNCE objective~\cite{chen2020simple}, this alignment loss encourages each query embedding to align closely with its correct class prototype (positive) while contrasting against all other prototypes (negatives). Unlike standard InfoNCE, which requires large batch sizes or memory banks to generate meaningful negatives, our formulation naturally provides structured, class-level negatives by leveraging the prototypes themselves as contrasting anchors. Since each prototype serves as a learned centroid for its class, this approach reduces reliance on instance-level diversity and yields stable gradients, making it particularly well-suited for low-shot regimes. 
We formalize this objective by computing temperature-scaled dot product similarities between each query and all class prototypes, followed by a softmax normalization to obtain class probabilities. Given a query embedding \( q_i \) and a set of class prototypes \( \{p_j\} \), the similarity score is defined as $s_{ij} = \frac{\langle q_i, p_j \rangle}{\tau}$, where \( \tau \) is a temperature hyperparameter that controls the sharpness of the similarity distribution. 
The alignment loss is defined as 

\begin{equation}
\mathcal{L}_{\text{align}} = -\sum_i \log \frac{\exp(s_{i y_i})}{\sum_j \exp(s_{ij})}
\end{equation}

\noindent This loss enforces discriminative alignment by maximizing the similarity between queries and their correct prototypes while minimizing similarity to all others, effectively shaping a robust and generalizable embedding space.
Together, these two objectives refine the embedding space beyond alignment, promoting both consistency across classification heads and geometric separation across classes. The overall training objective combines these terms with the prototype matching loss to compute the energy from Equation~\ref{eq:energy}.

\textbf{Learning.} 
The full model is trained in two successive stages, guided by the structured energy-based objective defined in Equation~\ref{eq:energy}, which unifies localization, classification, and embedding refinement through four components: the class-agnostic detection loss \( E_{\text{box}} \), prototype matching loss \( E_{\text{match}} \), classifier regularization via KL divergence \( D_{\text{KL}} \), and alignment-based separation loss \( E_{\text{align}} \). These energy terms correspond directly to the learning objectives used during training and are instantiated as the losses defined in Equation~\ref{eqn:loss_total}. 
In the first stage, we optimize the prototype matching loss \( \mathcal{L}_{\text{match}} \) to minimize energy between query embeddings and their corresponding class prototypes, encouraging stable alignment in the embedding space. We then refine this space by incorporating a distillation loss \( \mathcal{L}_{\text{KL}} \) to align classifier predictions with prototype-based outputs, and an alignment-based contrastive loss \( \mathcal{L}_{\text{align}} \) to enforce inter-class separation. The resulting training objective is:

\begin{equation}
\mathcal{L}_{\text{total}} = \mathcal{L}_{\text{match}} + \lambda_{\text{KL}} \mathcal{L}_{\text{KL}} + \lambda_{\text{align}} \mathcal{L}_{\text{align}}
\label{eqn:loss_total}
\end{equation}

\noindent where \( \lambda_{\text{KL}} \) and \( \lambda_{\text{align}} \) are scalar hyperparameters that weight the contribution of each loss term. During the first stage, we set \( \lambda_{\text{KL}} = \lambda_{\text{align}} = 0 \), and activate them in the second stage by setting both to 1. This staged optimization strategy shapes the overall energy landscape such that query embeddings are pulled toward their correct prototypes while being pushed away from incorrect or ambiguous ones, yielding a robust and discriminative embedding space that supports few-shot classification and open-set recognition.

\begin{table}[t]
\centering
\begin{tabular}{cc|c|c|c}
\hline
\multirow{2}{*}{\textbf{Approach}} & \multicolumn{2}{c|}{\textbf{5 Samples}} & \multicolumn{2}{c}{\textbf{10 Samples}} \\
\cmidrule{2-5}
 & \textbf{mAP} & \textbf{mAR} & \textbf{mAP} & \textbf{mAR} \\
\hline
DETR~\cite{carion2020end}               & 0.026  & 0.063  & \underline{0.054}  & 0.120 \\
YOLO~\cite{jocher2020ultralytics}               & 0.000  & 0.004  & 0.021  & \textbf{0.228} \\
\midrule
ProtoNet~\cite{snell2017prototypical} + SS      & 0.036  & \underline{0.119}  & 0.045  & 0.128 \\
ProtoKD~\cite{trehan2023protokd} + SS       & \underline{0.039}  & 0.116  & 0.048  & 0.142 \\
\midrule
ProtoKD~\cite{snell2017prototypical} + DETR      & 0.003  & 0.02  & 0.008  & 0.031 \\
ProtoNet~\cite{trehan2023protokd} + DETR       & 0.007  & 0.036  & 0.008  & 0.046 \\
\midrule
Stable-DINO~\cite{liu2023detection} & 0.002 & 0.002 & 0.011 & 0.003\\
FCOS\cite{tian2019fcos} & 0.001 & 0.002 & 0.016 & 0.078 \\
{FSP-DETR (M)}  & \textbf{0.069} & 0.170 & 0.080 & 0.203 \\
{FSP-DETR}  & 0.066 & \textbf{0.171} & \textbf{0.094} & \underline{0.207} \\

\hline
\end{tabular}
\caption{\textbf{Few-shot Ova Detection Results.} We report mean Average Precision (mAP) and mean Average Recall (mAR) for 5-shot and 10-shot settings. 
FSP-DETR consistently outperforms other methods, especially in low-data regimes. SS: Selective Search.
}
\label{tab:fewshot_detection}
\end{table}

\textbf{Inference.}
At inference time, we retain the standard DETR decoding pipeline to predict a fixed set of bounding boxes for each test image. However, unlike traditional DETR-based detectors that rely on a fixed classification head, our approach performs classification by comparing each detected box's feature embedding against a set of class prototypes. 
These prototypes can be constructed from any small support set, allowing us to perform few-shot classification on previously unseen ova categories by simply averaging the features of a few labeled exemplars. This flexibility enables our model to handle both standard test-time classification on known classes and generalization to novel classes without retraining. 
To reduce false positives, we include a background prototype derived from non-matching boxes during training. During inference, if a predicted box is closer to the background prototype than to any known class prototype, it is treated as a spurious detection and rejected. Moreover, we observe that averaging prototypes across semantically similar known classes can help identify coarse-grained categories or unknown ova types, supporting a soft form of open-set recognition. 
This prototype-driven inference mechanism allows for highly data-efficient adaptation and robust detection, even with sparse supervision.

\textbf{Implementation Details.} 
We implement our approach in PyTorch using DETR~\cite{carion2020end} with MS-COCO pretrained weights. The detector is fine-tuned in a binary setting for ova detection over 100 epochs using AdamW (lr = 1e-4, weight decay = 1e-4), with all foreground classes collapsed into one. We extract decoder features for both support and query boxes, passing them through a two-layer MLP (hidden dim = 512, ReLU) to obtain embeddings. Prototypes are computed from 5 annotated exemplars per class; queries are generated using the albumentations library with random flips, rotations, noise, and color perturbations. Training proceeds in two stages: we first optimize the prototype matching loss, then activate KL and alignment losses with weights \( \lambda_{\text{KL}} = \lambda_{\text{align}} {=} 1 \) and temperature \( \tau {=} 10 \). All experiments are run on a workstation with an NVIDIA A5500 GPU. 

\begin{table}[t]
\centering
\resizebox{\columnwidth}{!}{
\begin{tabular}{cc|c|c|c|c|c|c|c}
\hline
\multirow{2}{*}{\textbf{Approach}} & \multicolumn{2}{c|}{\textbf{FSP-DETR}} 
 & \multicolumn{2}{c|}{\textbf{FSP-DETR (M)}}
 & \multicolumn{2}{c|}{\textbf{SS+ProtoKD}} & \multicolumn{2}{c}{\textbf{SS+ProtoNet}} \\
\cmidrule{2-9}
 & \textbf{mAP} & \textbf{mAR} & \textbf{mAP} & \textbf{mAR} & 
 \textbf{mAP} & \textbf{mAR} & \textbf{mAP} & \textbf{mAR} \\
\midrule
Known     & \textbf{0.05} & 
{0.24} & 
\textbf{0.05} & \textbf{0.32} &
\underline{0.03} & \underline{0.31} & 0.00 & 0.08\\
Unknown   & \textbf{0.02} & \textbf{0.74} 
& \textbf{0.02} & 0.47 
& 0.00 & 0.02 & \underline{0.00} & \underline{0.04}\\
\bottomrule
\end{tabular}
}
\caption{\textbf{Open-set Ova Detection Results.} 
We evaluate detection performance across 15 known ova species and an \textit{unknown} class for unseen species. 
The unknown prototype is constructed by averaging seen class prototypes without access to unknown examples. 
``FSP-DETR (M)'': our model trained only with the matching loss.
}
\label{tab:openset_results}
\end{table}

\section{Quantitative Evaluation}\label{sec:eval}

\textbf{Data.}
We assembled a dataset comprising 3,001 ova examples, curated from 704 clinical samples collected at a local diagnostics laboratory (affiliation redacted for anonymity).
Each ovum type was isolated using centrifugal fecal flotation from samples of various hosts infected by distinct parasitic species. Identification and verification were conducted by a certified parasitologist through microscopic examination. Subsequently, high-resolution images were captured utilizing an Olympus BX43 microscope (Olympus Corporation, Tokyo, Japan) in conjunction with Olympus CellSens Entry software v1.18. To ensure clinical relevance, we selected 20 parasitic ova types based on their prevalence in clinical submissions received at both local and national diagnostic facilities. These ova include: Capillarids, Cystoisospora spp., Dipylidium caninum, Eimeria spp., Giardia spp., Moniezia sp., Nematodirus sp., Parascaris sp., Strongyles, Taeniid eggs. Toxascaris leonina, Toxocara spp., Trichostrongyles, Trichuris spp., Ancylostoma sp., Heterobilharzia americana, Eimeria macusaniensis, Eimeria leuckarti, Baylisascaris procyonis and Alaria sp.

\noindent \textbf{Baselines and Metrics}
To benchmark the effectiveness of our method, we compare it against both standard object detectors and prototype-based few-shot baselines under consistent low-shot conditions. Specifically, we fine-tune DETR~\cite{carion2020end}, YOLOv9~\cite{jocher2020ultralytics}, FCOS~\cite{tian2019fcos}, and Stable-DINO~\cite{Liu2024GroundingDINO} using 5 or 10 annotated bounding boxes per class across 15 seen ova categories. We also include FSP-DETR (M), a variant of our model trained with only the matching loss, to isolate the impact of our full objective. Additionally, we evaluate ProtoNet+SS and ProtoKD+SS~\cite{trehan2023protokd}, which rely on handcrafted region proposals via selective search over morphologically preprocessed images, followed by prototype-based classification. 
All models are evaluated using mean average precision (mAP) and recall (mAR) across IoU thresholds 0.50–0.95~\cite{lin2014microsoft,COCOAPI}. 




\begin{table}[t]
\centering
\small
\begin{tabular}{lcc}
\toprule
\textbf{Method} & \textbf{Precision (mAP)} & \textbf{Recall (mAR)} \\
\midrule
\multicolumn{3}{c}{\textit{(A) Unseen-Only (UO)}} \\
\midrule
ProtoNet + SS          & 0.049 & \underline{0.215}  \\
ProtoKD + SS           & \underline{0.053} & 0.213  \\
FSP-DETR (M) & 0.051 & 0.220\\
FSP-DETR     & \textbf{0.063} & \textbf{0.259}  \\

\midrule
\multicolumn{3}{c}{\textit{(B) Mixed-Prototypes, Unseen-Eval (MP-U)}} \\
\midrule
ProtoNet + SS          & 0.034 & 0.171 \\
ProtoKD + SS           & \underline{0.052} & \underline{0.196} \\
FSP-DETR (M) & 0.046 & 0.189\\
FSP-DETR     & \textbf{0.057} & \textbf{0.214} \\
\midrule
\multicolumn{3}{c}{\textit{(C) Mixed-Prototypes, Seen-Eval (MP-S)}} \\
\midrule
ProtoNet + SS          & 0.003 & 0.024 \\
ProtoKD + SS           & \underline{0.038} & \underline{0.111} \\
FSP-DETR (M) & 0.017 & 0.082\\
FSP-DETR     & \textbf{0.057} & \textbf{0.117} \\

\bottomrule
\end{tabular}
\caption{\textbf{Zero-shot Detection Results} across three evaluation settings: (A) \textit{Unseen-Only (UO)}, (B) \textit{Mixed-Prototypes, Unseen-Eval (MP-U)}, and (C) \textit{Mixed-Prototypes, Seen-Eval (MP-S)}. 
}
\label{tab:zero_shot_detection}
\end{table}

\subsection{Few-shot Ova Detection}
Table~\ref{tab:fewshot_detection} compares FSP-DETR against both parametric and prototype-based baselines under 5-shot and 10-shot settings. Our method achieves the best overall performance, particularly in the low-data regime (mAP 0.066 / mAR 0.171 at 5-shot), outperforming traditional detectors like DETR and YOLOv9 that struggle with sparse supervision. Although YOLO attains the highest recall at 10-shot (mAR 0.228), its low precision (mAP 0.021) suggests over-detection and poor background discrimination. Prototype-based methods using Selective Search (ProtoNet+SS, ProtoKD+SS) offer better recall than DETR/YOLO at 5-shot but are computationally inefficient and lack end-to-end learning. Recent backbones such as Stable-DINO and FCOS also perform poorly, indicating limited adaptability in few-shot regimes without class-specific training. FSP-DETR bridges this gap by combining a class-agnostic DETR backbone with prototype-guided embedding refinement, achieving the highest mAP and mAR in both 5-shot (0.066/0.171) and 10-shot (0.094/0.207) settings. While YOLO slightly surpasses FSP-DETR in recall at 10-shot, its much lower precision highlights its inferior discriminative capability. These results validate FSP-DETR’s design: integrating detection and classification through prototype-based supervision yields robust few-shot generalization. 

\subsection{Open-set Recognition}
We evaluate open-set ova detection by introducing an additional “unknown” class that aggregates novel species absent from training. This mimics real-world diagnostic settings where rare or emergent species may appear. To detect such unseen classes, we construct an unknown prototype by averaging seen and background class prototypes, without any access to novel class examples. As shown in Table~\ref{tab:openset_results}, {FSP-DETR} demonstrates strong generalization, achieving the highest recall on unknown classes (mAR=0.74) while maintaining competitive precision (mAP=0.02), indicating its ability to detect novelty from learned semantics. It also performs well on known categories (mAP=0.05, mAR=0.24). In contrast, SS-based baselines struggle to balance seen-unseen tradeoffs: while SS+ProtoKD reaches comparable mAR on known classes (0.31), it fails to detect unknowns (mAR=0.02, mAP=0.00). SS+ProtoNet underperforms on both fronts. Notably, even the simplified variant {FSP-DETR (M)}, trained without alignment or KL losses, achieves improved recall on known (mAR=0.32) and unknown (mAR=0.47) categories over prior baselines, supporting FSP-DETR's robustness under limited supervision. 
\subsection{Zero-shot Generalization}

\begin{table}[t]
\centering
\begin{tabular}{cc|c|c|c}
\hline
\multirow{2}{*}{\textbf{Approach}} & \multicolumn{2}{c|}{\textbf{Blood Cell}} & \multicolumn{2}{c}{\textbf{Malaria-infection}} \\
\cmidrule{2-5}
 & mAP & mAR & mAP & mAR \\
\hline
DeTR              & 0.001 & 0.041 & 0.007 & 0.076 \\
ProtoNet + SS     & 0.002 & 0.010 & 0.006 & 0.040 \\
ProtoKD + SS      & 0.004 & 0.018 & 0.002 & 0.034 \\
\textbf{FSP-DETR} & \textbf{0.017} & \textbf{0.048} & \textbf{0.010} &\textbf{0.088} \\
\hline
\end{tabular}
\caption{\textbf{5-shot Performance on other biomedical datasets:} \textit{Blood Cell Detection} (detecting three common cell types) and \textit{Malaria Detection} (detecting six uninfected and infected cells). 
}
\label{tab:biomedical_generalization}
\end{table}

To assess the generalization capacity of FSP-DETR, we evaluate it under three zero-shot detection scenarios shown in Table~\ref{tab:zero_shot_detection}. In the Unseen-Only (UO) setting, where both prototypes and evaluation are restricted to five entirely novel ova classes, FSP-DETR achieves the best performance (mAP=0.063, mAR=0.259), outperforming all baselines including FSP-DETR (M) (0.051 / 0.220), ProtoKD+SS (0.053 / 0.213), and ProtoNet+SS (0.049 / 0.215). This highlights the importance of our full training strategy, combining matching and detection losses, for effective zero-shot localization without prior exposure. 
In the Mixed-Prototypes, Unseen-Eval (MP-U) setting, where prototypes for both seen and unseen classes are included but evaluation is on the unseen classes, FSP-DETR again leads (0.057 / 0.214), with FSP-DETR (M) trailing slightly (0.046 / 0.189). In contrast, ProtoKD and ProtoNet exhibit degraded recall (${<}0.196$), showing their vulnerability to prototype interference. 
In the Mixed-Prototypes, Seen-Eval (MP-S) setting, which measures robustness to unseen prototypes while evaluating on seen classes, FSP-DETR maintains strong performance (0.057 / 0.117), significantly outperforming FSP-DETR (M) (0.017 / 0.082) and both prototype-based baselines (e.g., ProtoNet+SS: 0.003 / 0.024). This drop for FSP-DETR (M) suggests that while matching loss enables basic generalization, it is insufficient for disentangling seen and unseen prototypes during joint inference. 

\begin{table}[t]
\centering
\begin{tabular}{lcc}
\toprule
\textbf{Model} & \textbf{mAP} & \textbf{mAR} \\
\midrule
\multicolumn{3}{c}{\textit{Effect of embedding architecture}} \\
\midrule
DETR & 0.026 & 0.063 \\
FSP-DETR (2-layer MLP) & \textbf{0.066} & \textbf{0.171} \\
FSP-DETR (3-layer MLP) & 0.062 & 0.153 \\
FSP-DETR (4-layer MLP) & 0.056 & 0.138 \\
\midrule
\multicolumn{3}{c}{\textit{Effect of different loss functions}} \\
\midrule
$\mathcal{L}_{\text{Matching}}$ & \textbf{0.069} & 0.170 \\
$\mathcal{L}_{\text{Matching}} + \mathcal{L}_{\text{KL}}$ & 0.012 & 0.091 \\
$\mathcal{L}_{\text{Matching}} + \mathcal{L}_{\text{align}}$ & 0.052 & {0.169} \\
$\mathcal{L}_{\text{Matching}} + \mathcal{L}_{\text{KL}} + \mathcal{L}_{\text{align}}$ & 0.066 & \textbf{0.171} \\
\midrule
\multicolumn{3}{c}{\textit{Effect of support-query split and augmentation}}\\
\midrule
Partial Split (W/ aug) & \textbf{0.071} & 0.161 \\
Partial Split (No aug) & 0.035 & 0.071 \\
\midrule
\multicolumn{3}{c}{\textit{Effect of Temperature}}\\
\midrule
$\tau=0.1$  & 0.066 & 0.151 \\
$\tau=0.5$  & \textbf{0.069} & 0.164 \\
$\tau=1$    & 0.067 & 0.147 \\
$\tau=10$   & {0.066} & \textbf{0.171} \\
\bottomrule
\end{tabular}
\caption{\textbf{Ablation studies} to evaluate the effect of embedding architectures, loss functions, and using labeled samples for both support and query. “Partial Split” denotes 3/5 images used as support and 2/5 as query; “w/ aug”: support includes augmented images.}
\label{tab:ablation}
\end{table}

\begin{figure*}[ht]
    \centering
    \begin{tabular}{ccc}
        \textbf{Few-shot} & \textbf{Zero-shot} & \textbf{Open-set} \\
        \includegraphics[width=0.245\linewidth]{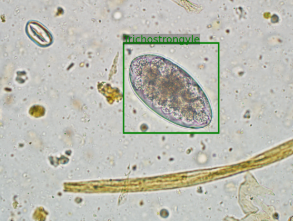} &
        \includegraphics[width=0.245\linewidth]{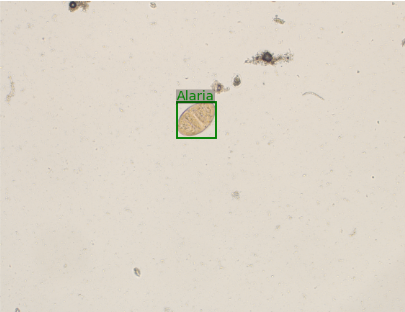} &
        \includegraphics[width=0.245\linewidth]{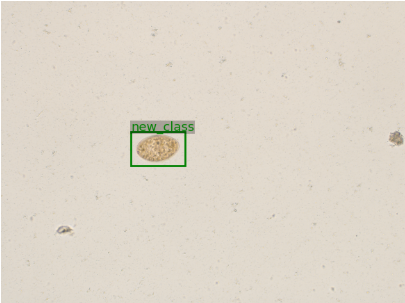} \\
        \includegraphics[width=0.245\linewidth]{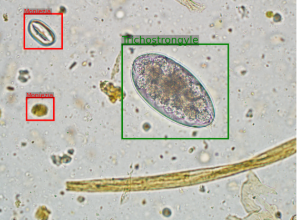} &
        \includegraphics[width=0.245\linewidth]{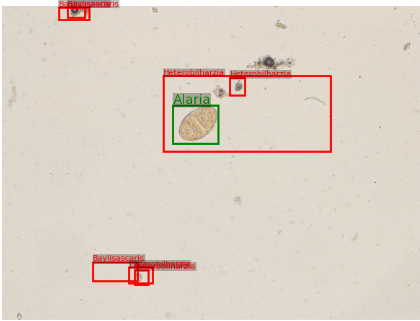} &
        \includegraphics[width=0.245\linewidth]{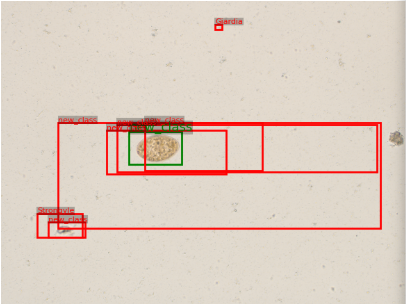} \\
        \multicolumn{3}{c}{
            \small Ground truth on top row; predictions below (green boxes = true positives)
        }
    \end{tabular}
    \caption{\textbf{Qualitative Visualization} of FSP-DETR's performance across three tasks: few-shot, zero-shot, and open-set ova detection. Each column shows a representative example, with the top row showing ground truth and the bottom showing predicted detections.}
    \label{fig:qual_viz}
\end{figure*}

\textbf{Generalization to Other Domains}
To evaluate cross-domain generalization under limited supervision, we test our model on two other biomedical detection tasks with 5 samples per class. The \textit{Blood Cell Detection} task~\cite{BCCD} involves identifying red blood cells (RBCs), white blood cells (WBCs), and platelets in microscopy images. The \textit{Malaria Detection} task~\cite{ljosa2012annotated} comprises six fine-grained classes: two uninfected cell types (RBCs and leukocytes) and four infected types (gametocytes, rings, trophozoites, and schizonts). Annotators were also allowed to label ambiguous cases as \textit{difficult}, increasing the complexity of the recognition task. In our evaluation, we ignore the ambiguous case and focus on the six distinguishable classes. This evaluation assesses whether the framework can generalize effectively to unseen biomedical domains with varying visual statistics, class granularity, and annotation ambiguities. As shown in Table~\ref{tab:biomedical_generalization}, {FSP-DETR} achieves the highest performance across both tasks, obtaining the best mAP and mAR on Blood Cell Detection (0.017 and 0.048, respectively) and on Malaria Detection (0.010 and 0.088, respectively). 
FSP-DETR maintains both higher precision and recall, demonstrating robustness to clutter and inter-class similarity. 



\textbf{Ablation Studies.} 
Table~\ref{tab:ablation} presents a comprehensive ablation of our framework, focusing on the effect of embedding architecture, loss functions, support-query splits, and temperature. A 2-layer MLP yields the best performance (mAP: 0.066, mAR: 0.171), outperforming both deeper variants and the DETR baseline, indicating that moderate capacity balances expressivity and generalization in low-shot settings. Among loss configurations, using only the matching loss achieves the highest precision (mAP: 0.069), while combining it with KL and alignment losses offers the best recall (mAR: 0.171) without degrading mAP, highlighting complementary benefits: matching ensures alignment, KL promotes calibration, and alignment encourages separation. Splitting the support/query set and augmenting support images boosts both metrics (mAP: 0.071, mAR: 0.161), whereas training without augmentation suffers notably. Lastly, varying the temperature $\tau$ shows mild impact, with values between 0.1–10 yielding comparable results, and $\tau = 10$ marginally outperforms others in recall. 

\textbf{Qualitative Discussion.} 
We provide qualitative results in Figure~\ref{fig:qual_viz} across three challenging settings: few-shot detection (top row), zero-shot detection (middle row), and open-set recognition (bottom row). In each pair, the left image shows ground truth annotations (green boxes), while the right shows model predictions, with true positives in green and false positives/negatives in red. In the few-shot setting, FSP-DETR accurately localizes and classifies instances with only five support samples per class. In the zero-shot case, the model can detect ova instances of the unseen Alaria sp. class using a prototype constructed solely from a visual example. In the open-set setting, five unseen classes are modeled as a single “new class” prototype during training. Despite the increased difficulty, FSP-DETR successfully localizes novel class instances and demonstrates its capacity to generalize and reject unseen classes. 


\section{Conclusion and Future Work}\label{sec:conclusion}
In this work, we introduced FSP-DETR, a prototype-driven framework for few-shot ova detection that integrates DETR-based object proposals with trainable embeddings and joint prototype alignment. We demonstrate strong performance across low-shot, open-set, and generalization to biomedical detection tasks, consistently surpassing competitive baselines. Despite these gains, several limitations remain. The current use of prototype averaging for unseen class detection may oversimplify the semantic diversity and limit representational expressiveness. Similarly, the fixed MLP projection head, while effective, may not fully capture complex visual correspondences in heterogeneous biomedical data. Using limited examples for prototype construction makes the model sensitive to noisy or low-quality image samples. In the future, we aim to address these limitations by exploring adaptive embedding mechanisms for uncertainty-aware learning for robust learning. 

\textbf{Acknowledgements.} This work was supported by the US NSF grants IIS 2348689 and IIS 2348690, and the USDA award no. 2023-69014-39716.

{
    \small
    \bibliographystyle{ieeenat_fullname}
    \bibliography{main}
}

\end{document}